\documentclass[letterpaper, 10 pt, conference]{ieeeconf}  

\IEEEoverridecommandlockouts                              
\usepackage{graphicx} 
\usepackage{bm}
\usepackage{amsmath}
\usepackage{xcolor}
\usepackage{amssymb} 
\usepackage{mathtools} 

\usepackage{amsthm}
\usepackage{hyperref}
\usepackage{pgfplots}
\pgfplotsset{compat=1.18}
\usepackage{xcolor} 
\usepackage{graphicx} 
\usepackage{caption}  
\usepackage{epstopdf}
\usepackage{textcomp}
\usepackage{gensymb}
\usepackage{subfigure} 
\usepackage{subcaption} 
\usepackage{dblfloatfix}

\theoremstyle{definition}

\title{\LARGE \bf
Learning to Nudge: A Scalable Barrier Function Framework for Safe Robot Interaction in Dense Clutter
}

\author{
Haixin Jin$^{*,1}$, Nikhil Uday Shinde$^{*,1}$, Soofiyan Atar$^{1}$, Hongzhan Yu$^{1}$, Dylan Hirsch$^{1}$,\\ Sicun Gao$^{1}$, Michael C. Yip$^{1}$, Sylvia Herbert$^{1}$\\
$^{1}$University of California San Diego\\
{\tt\small \{haj013, nshinde, satar, hoy021, dhirsch, sig049, yip, sherbert\}@ucsd.edu}
\thanks{$^{*}$These authors contributed equally to this work.}
}

\begin{document}

\maketitle
\thispagestyle{empty}
\pagestyle{empty}


\newcommand{\ourcbf}{Dense Contact Barrier Functions}
\newcommand{\ourcbfAcronym}{DCBF}

\newcommand{\R}{{R}}
\newcommand{\action}{u} 
\newcommand{\actionT}{u^{t}} 
\newcommand{\actionTpone}{u^{t+1}} 
\newcommand{\actionTmone}{u^{t-1}} 
\newcommand{\state}{\mathbf{r}}            
\newcommand{\stateT}{\mathbf{r}^{t}} 
\newcommand{\stateTpone}{\mathbf{r}^{t+1}} 
\newcommand{\stateTmone}{\mathbf{r}^{t-1}} 
\newcommand{\objstate}{\mathbf{o}}         
\newcommand{\objstateT}{\mathbf{o}^{t}} 
\newcommand{\objstateTpone}{\mathbf{o}^{t+1}} 
\newcommand{\objstateTmone}{\mathbf{o}^{t-1}} 
\newcommand{\objstateHist}{\mathbf{O}}         
\newcommand{\objstateHistT}{\mathbf{O}^{t}} 
\newcommand{\objstateHistTpone}{\mathbf{o}^{t+1}} 
\newcommand{\objstateHistTmone}{\mathbf{O}^{t-1}} 
\newcommand{\objDynamicsError}{e} 
\newcommand{\superlevelset}{\sigma^*}

\newcommand{\safetyFn}{q}    
\newcommand{\safetyThresh}{q_{T}} 
\newcommand{\hist}{h}                      
\newcommand{\Hspace}{\mathcal{H}}          
\newcommand{\B}{B}                         
\newcommand{\Bglobal}{B_{\mathrm{global}}} 
\newcommand{\C}{\mathcal{C}}               
\newcommand{\classK}{\alpha}               
\newcommand{\hinge}{\phi_\gamma}           
\newcommand{\thetathr}{\theta_{\max}}      
\newcommand{\Nobj}{N}       
\newcommand{\Dsafe}{\mathcal{D}_s}         
\newcommand{\Dunsafe}{\mathcal{D}_u}       
\newcommand{\Dall}{\mathcal{D}}            
\newcommand{\uNom}{u_{\mathrm{nom}}}       
\newcommand{\uSafe}{u_{\mathrm{safe}}}     
\newcommand{\nexts}[1]{#1^{+}}             
\newcommand{\Ns}{N_s}                      
\newcommand{\Nu}{N_u}                      
\newcommand{\Nall}{N}                      

\begin{abstract}
Robots operating in everyday environments must navigate and manipulate within densely cluttered spaces, where physical contact with surrounding objects is unavoidable. 
Traditional safety frameworks treat contact as unsafe, restricting robots to collision avoidance and limiting their ability to function in dense, everyday settings. 
As the number of objects grows, model-based approaches for safe manipulation become computationally intractable; meanwhile, learned methods typically tie safety to the task at hand, making them hard to transfer to new tasks without retraining.
In this work we introduce \ourcbf (\ourcbfAcronym). 
Our approach bypasses the computational complexity of explicitly modeling multi-object dynamics by instead learning a composable, object-centric function that implicitly captures the safety constraints arising from physical interactions.
Trained offline on interactions with a few objects, the learned \ourcbfAcronym composes across arbitrary object sets at runtime, producing a single global safety filter that scales linearly and transfers across tasks without retraining. 
We validate our approach through simulated experiments in dense clutter, demonstrating its ability to enable collision-free navigation and safe, contact-rich interaction in suitable settings.


\end{abstract}


\section{Introduction}
The ability for robots to safely interact with and maneuver through densely cluttered environments is a critical step towards their widespread deployment in everyday, unstructured spaces. 
Such capabilities could enable household robots to safely navigate through crowded pantries, or allow warehouse robots to restock shelves in tight, shared-use commercial settings. 
These settings contain dense, dynamic arrangements of objects, often forcing the robot to make physical contact to complete its tasks. 
For example, a robot retrieving an item from a full pantry may need to gently push aside neighboring items, while a logistics robot may need to make contact with fragile goods without knocking them over. 

Research on robot safety has historically focused on collision avoidance. 
The majority of previous work from classical methods like Hamilton-Jacobi Reachability (HJR)~\cite{HJRSylvia}, Control Barrier Functions (CBFs)~\cite{CBFAmes} and Artificial Potential Fields (APF)~\cite{APF} are all predicated on maintaining some minimum distance from all obstacles. 
Even works focused on complex manipulation often handle safety by incorporating the manipulated object into the set of items with which collision must be avoided~\cite{DLOManipulation, safeClutteredManipulation}. 
However, this criterion of collision avoidance is fundamentally limiting in cluttered spaces. 
A home robot focused on strictly avoiding collisions would be paralyzed with the task of reaching any object beyond the first row of items in a pantry. 
Humans, in contrast, routinely make contact, nudging and shifting objects while implicitly understanding the interactions required to prevent spills or damage. 
This work aims to bridge this gap, giving robots the capacity to reason about and execute safe contact. 

While some research has explored physical interaction, these methods often fail to address the scalability and generalizability required for these dense multi-object scenarios. 
Approaches that explicitly model the complex, coupled dynamics between the robot and object~\cite{JIGGLE} become computationally intractable as the number of objects grows, due to the combinatorial explosion of potential interactions. 
Works such as~\cite{JIGGLE, latentken} successfully demonstrate safe contact with a single complex deformable object, but fail to provide a clear path to scaling to multi-object systems. 
Learned methods, such as those in reinforcement learning~\cite{oswinneurips, dohjpponeurips}, can handle complex systems but often entangle safety and performance, requiring complete retraining for each new task. 
This highlights the need for a method that balances performance with safety during these interactions, in a manner that scales to a large, variable number of objects and readily generalizes across diverse arrangements.

In this paper we introduce \ourcbf (\ourcbfAcronym), object-centric control barrier functions that compute implicit interactions to achieve safe, scalable interaction in dense object environments. 
The core of our framework is a composable \ourcbfAcronym \space that is trained offline on interactions with just a few objects. 
At runtime, this function is evaluated for each object in the scene, and the results are aggregated into a single safety filter. 
This compositional approach allows our method to scale to environments with an arbitrary number of objects and generalizes across diverse configurations without retraining. 
Our method avoids the intractability of explicit multi-object interaction modeling and reasoning by learning to implicitly capture the safety constraints arising from complex interactions in an object-centric, scalable manner. 
Our primary contributions are: 
\begin{itemize}
    \item An implicit interaction CBF formulation that captures the multi-object interaction effects on safety without explicit modeling. 
    \item A composable, object-centric framework that enables scalable safety reasoning for a variable number of objects. 
    \item A high-interaction refinement procedure to reduce the conservativeness of the learned CBF
    \item Validation of our approach in simulated experiments for safe interaction in densely cluttered scenes. 
\end{itemize}

\begin{figure*}[!t]
  \centering
  \includegraphics[trim=1.5cm 1.5cm 1.5cm 2.2cm, clip,width=\textwidth]{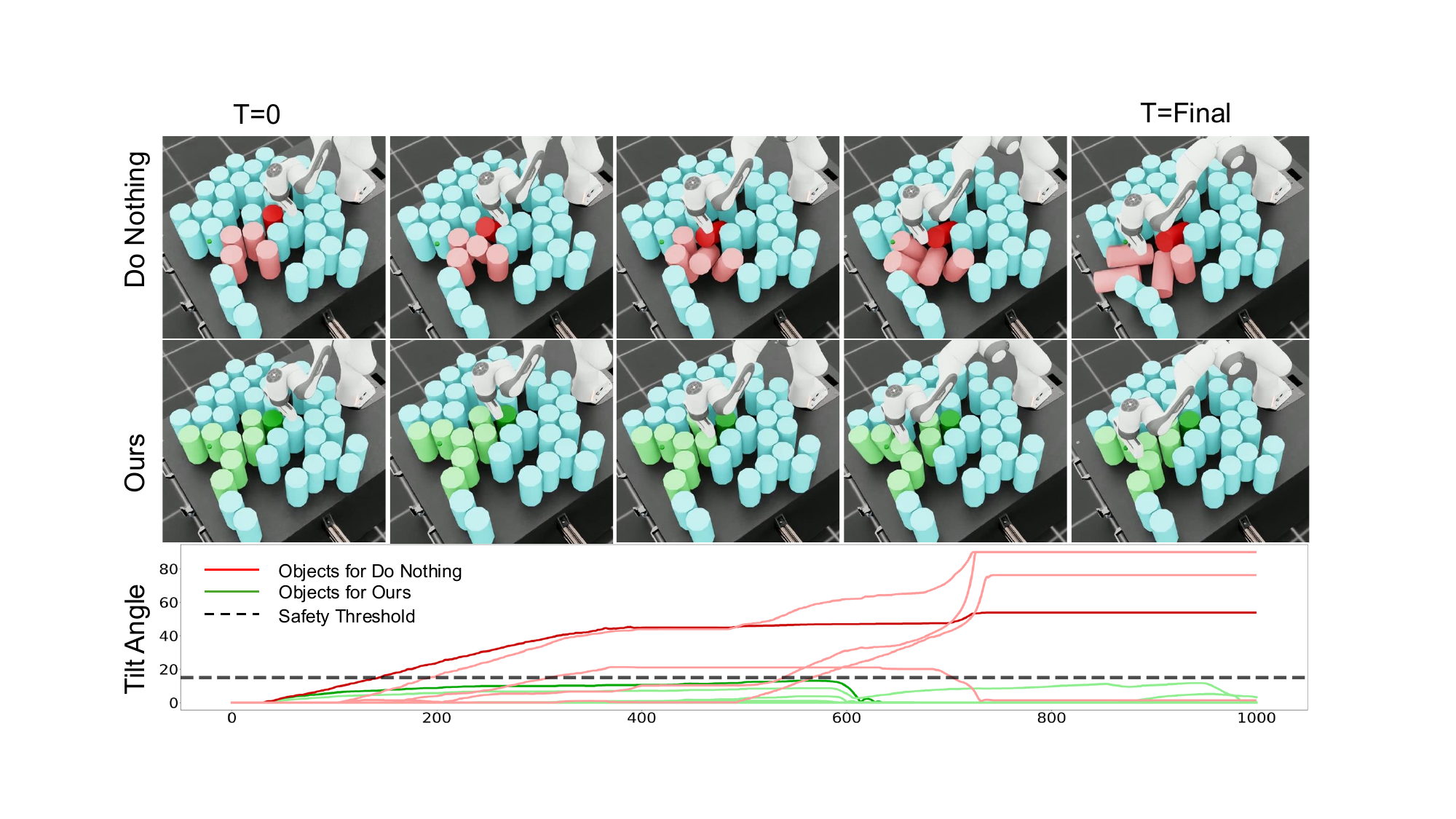}
  \caption{Simulated Robot trajectory from start to goal across different methods. 
The first row shows the \emph{Do Nothing} baseline, where the robot directly pushes through the clutter, causing multiple objects to be knocked over. Interacted objects are highlighted in red, with the object starting closest to the end-effector shown in dark red and others in light red. 
The second row shows our proposed method, which safely interacts with objects to reach the goal without violating safety. Interacted objects are highlighted in green, with the starting closest object shown in dark green and others in light green. 
The third row plots the tilt angles of the interacted objects over time, with curves matching the same color coding (dark/light red and dark/light green) with a dashed line denoting the safety threshold, $15\degree$. 
The red curves exceed tilt threshold, indicating safety violations, whereas the green curves remain below the threshold, demonstrating safe interactions.}

  \label{fig:trajPlot}
\end{figure*}

\section{Related Works}

The development of safe autonomous systems is a critical area of research, particularly for robots operating in complex dynamic environments such as households or commercial facilities.
While a large body of work exists on ensuring robot safety, much of it has centered on collision avoidance. 
This paradigm, though effective in open spaces, is overly restrictive in dense settings where robots must make physical contact and interact with objects to complete their tasks. 
Our work addresses this gap by proposing a scalable method for safe robot–object interaction through composable barrier functions and implicit interaction modeling.

\textbf{Safe Autonomy for Collision Avoidance: }
Robot safety has traditionally been synonymous with avoiding collisions. 
Classical methods like Hamilton-Jacobi Reachability (HJR) compute safety value functions that define control invariant safe sets~\cite{HJRSylvia}. 
These functions have been used online for optimal control and as minimally invasive safety filters~\cite{backtobase}. 
However, HJR suffers from the curse of dimensionality and requires analytical models, limiting its use to low dimensional, control affine systems and consequently simpler collision avoidance based safety.  
Recent works like~\cite{deepreach, deepreachmpc} extend HJR using learning-based approximations, alleviating dimensionality restrictions but they still depend on known analytic models, limiting their scalability to environments with unknown dynamic object interactions. 

This focus on collision avoidance persists even in manipulation tasks where object interaction is the goal~\cite{marcosafemanipulation}.
Classical methods such as potential fields~\cite{APF, APFHumanRobotCollab} and MPC-based planners~\cite{shortHorizonMPC} have been focused on maintaining separation from and minimal contact with obstacles.
More recent work in cluttered~\cite{safeClutteredManipulation} and deformable object manipulation~\cite{DLOManipulation} 
extends this notion of safe collision avoidance by treating the manipulated object as part of the system, thereby ensuring the object also avoids collisions with the environment.
These approaches remain overly conservative for crowded environments where avoiding all contact makes many tasks infeasible.

\textbf{Challenges in Scalability and Generalization: }
Methods that do explore more complex systems and even physical interaction often fail to scale to dense object environments. 
Reinforcement learning approaches have been used to handle complex dynamics, either by integrating safety as a soft constraint~\cite{cpo,gu2025balancerewardsafetyoptimization} or by using Bellman-inspired updates~\cite{oswinneurips,dohjpponeurips}. 
However, these methods typically entangle safety with task-specific policies, limiting their use as general, task-agnostic safety filters. 
Other approaches have leveraged advances in simulation technology such as XPBD~\cite{macklin2016xpbd} and IsaacLab~\cite{isaaclab} to enable planning for complex systems. 
For instance, \cite{JIGGLE} uses XPBD with model based planning to manipulate deformables under safety constraints for active sensing.
Other approaches like \cite{latentken} leverage learned latent space dynamics to reason about more complex definitions of safety in deformable manipulation tasks. 
Methods like~\cite{forcefeedback} even incorporate additional sensing modalities, like force feedback to safely manipulate a single articulated object. 
While effective for single-object settings, these methods remain computationally intensive or fail to generalize well to multi-object, cluttered environments. 
A key requirement is a scalable method that can reason about a variable number of movable objects without being overly conservative. 

\textbf{Control Barrier Functions: }
To address the challenges presented above, we build upon the framework of Control Barrier Functions (CBFs). 
A CBF is a function that defines a control invariant safe set in the system's state space, guaranteeing the existence of a control action that keeps the agent in this set~\cite{CBFAmes, discreteCBF}. 
CBFs have been widely used to enforce safety online as minimally invasive safety filters, with extensions such as Robust CBFs (RCBFs)~\cite{parameterizedCBF} generalizing them to uncertain dynamics.

A key challenge for classical CBFs is their synthesis~\cite{cbfchallenges}, as it often requires precise system knowledge. 
This limitation has historically restricted their use to simpler tasks such as collision avoidance. 
While HJR-based synthesis is possible~\cite{backtobase}, it suffers from poor scalability. 
To overcome this, Neural Control Barrier Functions (NCBFs) learn safety constraints directly from data~\cite{ncbf}, enabling a new class of methods that 
incorporate perception~\cite{RajaEtAl2024}, 
handle uncertainty~\cite{TayalEtAl2024}, and scale to dense environments~\cite{sncbf}. 
Other works continue to build upon this, enabling refining candidate NCBFs with HJR~\cite{refinecbf}. 
Works such as \cite{sncbf} have improved the scalability of safety in dense environments, leveraging composable CBFs to enable safe navigation in crowds. 
Despite this progress, the application of CBFs remains focused on collision avoidance, leaving a critical gap in methods that can manage safe physical interaction.

In this work, we tackle the problem of safe interaction in dense, multi-object environments where physical contact is unavoidable. 
We extend CBFs beyond collision avoidance by introducing \ourcbf which reason about per-object safety in a composable manner. 
This enables scalable and generalizable safety filters that can handle multiple objects in diverse configurations without sacrificing the ability to interact and complete tasks.

\begin{figure*}[!t]
  \centering
  \includegraphics[trim=1cm 6.5cm 2.8cm 3.5cm, clip,width=\textwidth]{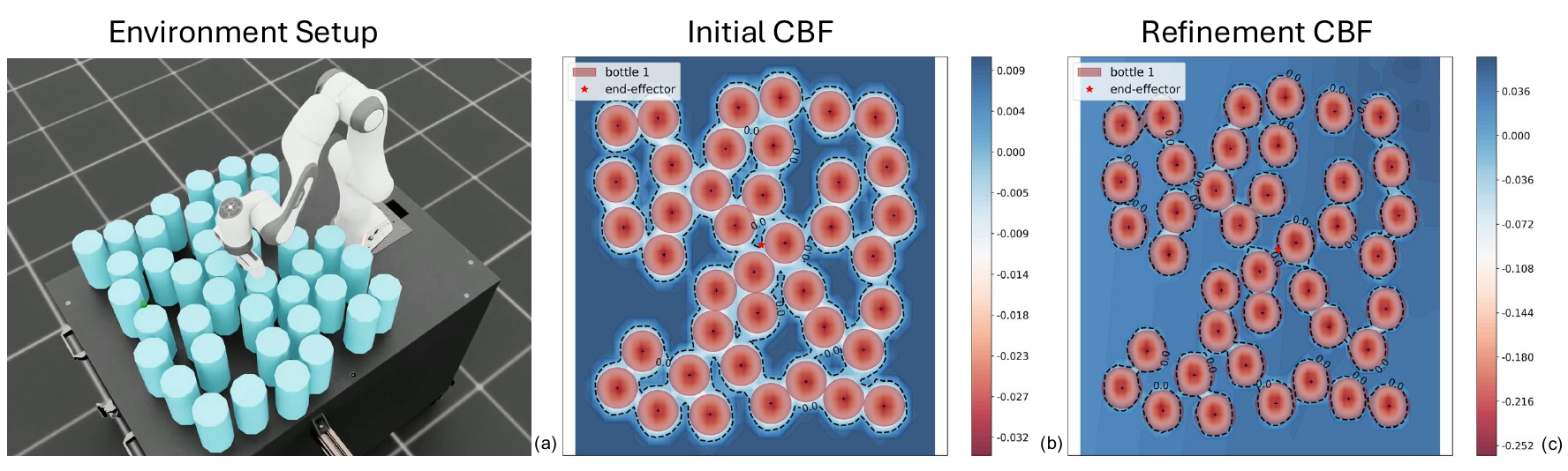}
  \caption{ Comparison of safe boundaries for the initial model and refined model ($\sigma=0.01$): (a) Environment snapshot with contact. The end-effector makes slight
contact with a bottle, pushing it by a small angle while keeping it within the safe region. (b) Global CBF value plot for the
initial model under contact. The conservative boundary incorrectly treats the interaction as unsafe, often leading to the robot
stalling. (c) Global CBF value plot for the refined model under contact. The boundary accommodates
safe interactions, reflecting 
less conservative, realistic reasoning about safety for dense environments}

  \label{fig:globalCBFPlot}
\end{figure*}

\section{Preliminaries}

\subsection{Problem Statement}

We consider the problem of a robot safely operating in environments densely populated with objects, such as a household pantry or the shelves of a store or warehouse. 
The robot may physically interact with the objects, but must ensure that its actions do not cause the objects to enter an unsafe state. 
Though our formulation generalizes to a variety of state-based safety constraints, in this paper we focus on the failure condition of objects tipping beyond a prescribed tilt angle. 
This problem statement captures a variety of relevant-tasks such as a manipulator reaching for a glass in the back of a pantry, a ground robot pushing aside furniture, or warehouse robots reorganizing shelves. 
In each case, safety is defined in terms of object stability, while the robot must remain effective at task execution. 

We begin by formalizing this problem. 
We define the state of the robot as $\state \in \mathbb{R}^{n}$ with dynamics 
\begin{equation}\label{eq:robotdynamics}
\stateTpone = f(\stateT, \actionT),
\end{equation}
 where $\actionT \in \mathbb{R}^{a}$ is the control input. 
The environment contains $\Nobj$ objects each with a state $\objstate_i \in \mathbb{R}^{m}~ \forall i \in [1, \Nobj]$. 
The safety of each object is defined by a function $\safetyFn_i(\objstate_i)$, where a violation occurs if this safety cost exceeds a known threshold, i.e. $\safetyFn_i(\objstate_i) \geq \safetyThresh$.

For a scene with a single object, $\objstate$, the dynamics are given by: $\objstate^{(t+1)} = p(\stateT, u^{(t)}, \objstate^{(t)})$. 
However for multiple objects the dynamics, given by
\begin{equation}\label{eq:objectdynamics}
\begin{aligned}
        \{\objstate_1^{(t+1)}, &\objstate_{2}^{(t+1)}, \dots, \objstate_{\Nobj}^{(t+1)} \}\\ &= F(\state^{(t)}, u^{(t)}, \{\objstate_1^{(t)}, \objstate_{2}^{(t)}, \dots, \objstate_{\Nobj}^{(t)} \}),
\end{aligned}
\end{equation}
are influenced by both the robot and inter-object interactions. 
A key challenge with reasoning over the interactions in $F$ is that the number of inter-object interactions scales exponentially with the number of objects. 
Reasoning about the safety of each object through explicit modeling quickly becomes infeasible. 
Additionally, physics models for contact for dense environments are often too computationally inefficient for use. 
In this paper we assume access to the forward dynamics model of the robot~\eqref{eq:robotdynamics}, but do not have runtime access to the physics model~\eqref{eq:objectdynamics}, that forward simulates the objects.

\subsection{Control Barrier Functions}

Consider a discrete-time system with dynamics $\mathbf{x}^{t+1} = f(\mathbf{x}^{t}, u^{t})$ and state space $\mathbf{x} \in \mathbb{R}^n$.
We say that a set $\mathcal{C} \subseteq \mathbb{R}^n$ is \textit{control-invariant} if for any $\mathbf{x} \in \mathbb{R}^n$, there is an action $u$ which keeps the subsequent state in the set, i.e. $f(\mathbf{x}, u) \in \mathcal{C}$.

Moreover, we say a function $h:\mathbb{R}^n \to \mathbb{R}$ is a (discrete-time) \textit{control barrier function} (CBF) if for each state $\mathbf{x} \in \mathbb{R}^n$, the condition
\begin{equation*}
    h(f(\mathbf{x},u)) - h(\mathbf{x}) \ge -\gamma(h(\mathbf{x}))
\end{equation*}
holds, where $\gamma:\mathbb{R} \to \mathbb{R}$ is any extended class $\mathcal{K}$ function.

We note that the super-level set $\mathcal{C}_h := \{\mathbf{x} \in \mathbb{R}^n \mid h(\mathbf{x}) \ge 0\}$ for a CBF $h$ is control-invariant.

\section{\ourcbf: }
We aim to implicitly encode the effects that the inter-object contact dynamics have on each object into the structure of a learned CBF $B(\stateTpone_i, \objstateHistT_i)$, where $\stateTpone_i$ and $\objstateHistT_i$ are the object-centric robot state and the object-centric history of the robot-object interaction corresponding to object $i$. This will result in a task-agnostic safety filter that allows the robot to make safe contact with the objects while moving through the environment. In this section, we will explain three critical intricacies of our learned CBF model: 1) implicit interactions, wherein we implicitly encode inter-object interactions into the CBF, 2) state history, which improves implicit knowledge of interaction dynamics, and 3) object-centric state representation, which allows for composition and generalizability to environments of varying densities.



\subsection{Implicit interaction} To explain how the robot-object interactions are implicitly encoded in  $B$, we first begin by considering the case of a single-object environment with robot dynamics~\eqref{eq:robotdynamics} and object dynamics $\objstateTpone = p(\objstateT, \stateT, \actionT)$.
As contact dynamics can be difficult to analytically model and computationally inefficient to simulate, 
we will need to learn a data-driven nominal contact model $ \objstateTpone \approx \hat{p}(\objstateT, \stateT, \actionT)$.
In other words,
\begin{align}
\label{eq:p_phat_ptilde}
    p(\objstateT, \stateT, \actionT) = \hat{p}(\objstateT, \stateT, \actionT) + \objDynamicsError(\objstateT, \stateT, \actionT).
\end{align}
where $\objDynamicsError$ represents the error between the true and learned models. 

Following \cite{parameterizedCBF}, we say a function $h(\stateT, \objstateT)$ is a \textit{robust CBF} with respect to the learned dynamics $\hat{p}$ if
for some $\sigma > 0$, the inequality
\begin{align*}
    \sup_{u^{t} \in \mathbb{R}^{m}} &h\big(f(\stateT, \actionT), \hat{p}(\objstateT, \stateT, \actionT)\big) - h(\stateT, \objstateT) -\sigma \\\nonumber 
    &\geq -\gamma(h(\stateT, \objstateT))
\end{align*}
holds for each $\stateT$ and $\objstateT$.
The parameter $\sigma$ is called the \textit{robustness margin}.
The corresponding \textit{robust feasible control set} is: 
\begin{align}
    &K_{\text{RCBF}}(\stateT, \objstateT) = \Big\{ u \in \mathbb{R}^{m} | \\\nonumber 
    &h\big(f(\stateT, \actionT), \hat{p}(\objstateT, \stateT, \actionT)\big) - h(\stateT, \objstateT) - \sigma \geq -\gamma(h(\stateT, \objstateT))\Big\}.
\end{align}

To understand the utility of a robust CBF, consider the following.
Suppose we have some robust CBF $h$ for which the model mismatch is bounded above by the robustness margin, i.e.
\begin{equation*}
    h\big(f(\stateT, \actionT), \hat{p}(\objstateT, \stateT, \actionT)\big) - h\big(f(\stateT, \actionT), p(\objstateT, \stateT, \actionT)\big) \le \sigma
\end{equation*}
for all $\stateT$, $\objstateT$, and $\actionT$.
It then follows that
\begin{align*}
    &\sup_{u^{t} \in \mathbb{R}^{m}} h\big(f(\stateT, \actionT), p(\objstateT, \stateT, \actionT)\big) - h(\stateT, \objstateT)
    \\
    &= 
    \sup_{u^{t} \in \mathbb{R}^m} \bigg[ \underbrace{h\big(f(\stateT, \actionT), \hat{p}(\objstateT, \stateT, \actionT)\big) - h(\stateT, \objstateT)}_{\text{nominal}}  \\\nonumber
    &-\underbrace{\Big(h\big(f(\stateT, \actionT), \hat{p}(\objstateT, \stateT, \actionT)\big) - h\big(f(\stateT, \actionT), p(\objstateT, \stateT, \actionT)\big)\Big)}_{\text{mismatch}} \bigg]\\
    &\ge\sup_{u^{t} \in \mathbb{R}^{m}} h\big(f(\stateT, \actionT), p(\objstateT, \stateT, \actionT)\big) - h(\stateT, \objstateT) - \sigma\\
    &\geq -\gamma(h(\stateT, \objstateT)).
\end{align*}
Thus if $h$ is a robust CBF with respect to the learned dynamics and the model mismatch is bounded above by the robustness margin, then $h$ is also a CBF with respect to the true dynamics.
In particular, any controller $u = k(\stateT, \objstateT) \in K_{\text{RCBF}}(\stateT, \objstateT)$ guarantees safety, not just under the learned dynamics, but under the true dynamics as well.

It is for this reason that we will seek to learn both the object dynamics and a robust CBF (although we will do so implicitly, as discussed shortly).
Further, note that by the mean-value theorem, $h(f(\stateT, \actionT), \hat{p}(\objstateT, \stateT, \actionT)) - h\big(f(\stateT, \actionT), p(\objstateT, \stateT, \actionT)\big) =- \frac{\partial h}{\partial \objstate}(f(\stateT, \actionT), \objstate^*) \objDynamicsError(\objstateT, \stateT, \actionT)$
for some $\objstate^*$ between $\hat{p}(\objstateT, \stateT, \actionT)$ and $p(\objstateT, \stateT, \actionT))$.
Given a Lipschitz bound $L_{o}$ on how the CBF value $h$ changes with the object positions, and an error bound $\varepsilon \ge \|e\|$, a sufficient robustness margin is given by
\begin{align*}
    \sigma^* &:= L_{0} \varepsilon \ge - \frac{\partial h}{\partial \objstate}(f(\stateT, \actionT), \objstate^*) \objDynamicsError(\objstateT, \stateT, \actionT)\\
    &\ge h(f(\stateT, \actionT), \hat{p}(\objstateT, \stateT, \actionT)) - h\big(f(\stateT, \actionT), p(\objstateT, \stateT, \actionT)\big).
\end{align*}
As the model $\hat{p}$ improves, $\|\objDynamicsError\|$ and thus $\sigma^*$ shrinks proportionally.
This conceptually shows that as training data becomes more representative of possible interactions,
enabling 
$\hat{p}$ to better capture the dynamics, 
this sufficient robustness margin $\sigma^*$ decreases.

\begin{figure}
    \centering
    \includegraphics[ width=\linewidth, height=6cm]{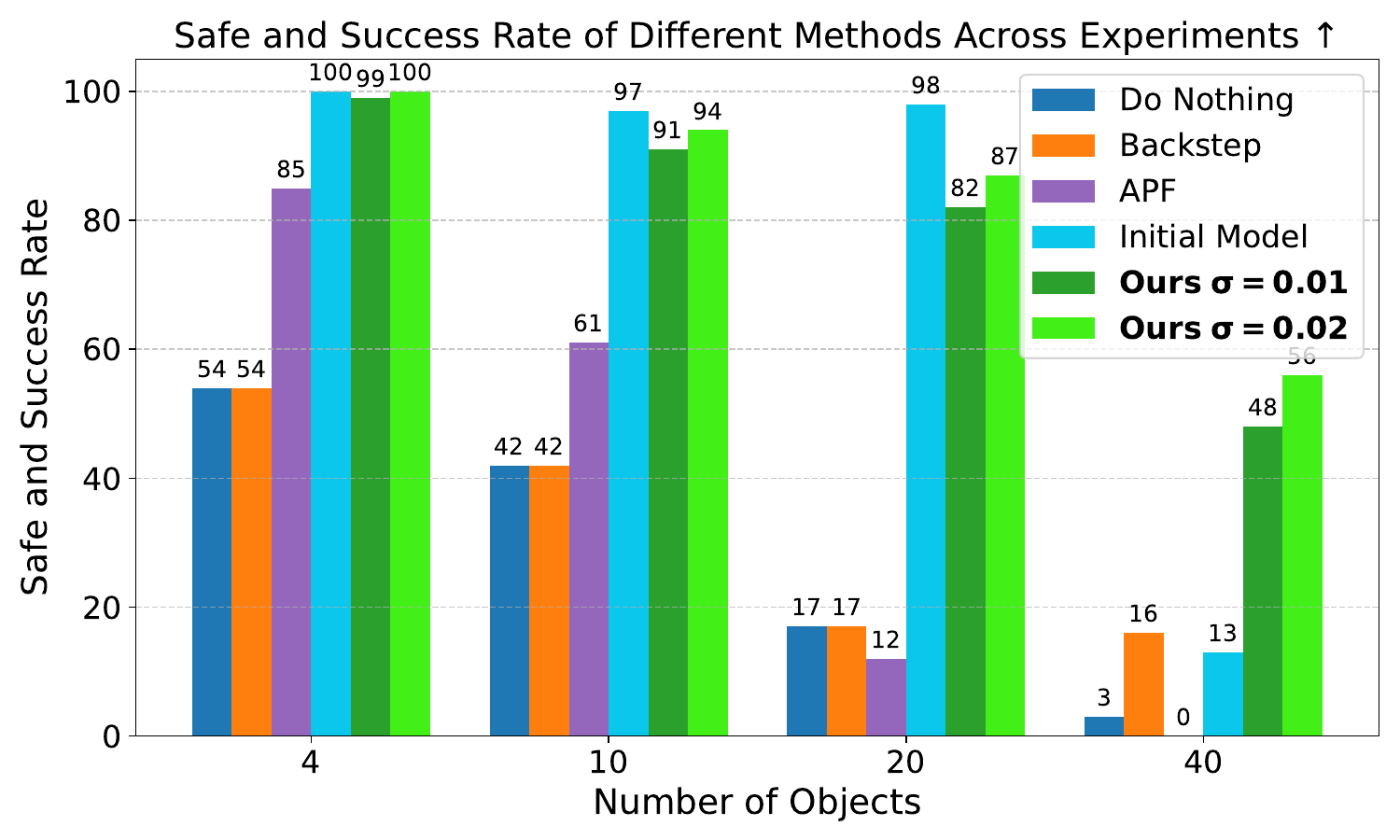}
    \caption{Safe and success rate of different methods evaluated across environments with increasing object density ($4$, $10$, $20$, and $40$ objects).
    This metric illustrates percentage of trajectories where the robot both reached the goal and stayed safe throughout the trajectory. 
    Results are computed over $100$ randomized trajectories for each method in each environment. 
    The results highlight the limitations of the baselines, especially at higher object densities and the robustness of our method in maintaining both safety and task success. 
    }
    \label{fig:safe_reach_success_rate}
\end{figure}

Finally, instead of learning a CBF $h$ and the dynamics $p$ explicitly and separately, we propose doing so implicitly and simultaneously by instead learning the function
$B(\stateTpone, \objstateT) \approx  h(\stateTpone, \objstateTpone)$.
Operationally, the function $B$ can be used at runtime as follows.
Suppose we have a robot-object state $(\stateT, \objstateT)$.
Before choosing an action $\actionT$, we determine what the next robot state will be under this action using the robot model, i.e. we evaluate $\stateTpone = f(\stateT, \actionT)$.
Next, we query the value of $B(\stateTpone, \objstateT) \approx h(\stateTpone, \objstateTpone)$ to determine the safety value of the chosen action.
If the corresponding robot-object state is safe, i.e. $B(\stateTpone, \objstateT) \ge 0$, we can safely take the chosen action.


\subsection{Object state history}
Inspired by~\cite{sncbf}, we modify the neural CBF model to take in not just the single-step object state $\objstateT$, but a history of the object states over the past $T$ time steps: $\objstateHistT_{\textbf{abs}} = [\objstateT, \objstateTmone, \dots, \objstate^{t-(T-1)}]$. 
We do so to better encode latent interaction effects (robot-object and object-object) and unobserved attributes (e.g. friction, object mass).


\subsection{Object-centric representation}
Implicit interaction modeling also readily generalizes to multiple objects. 
However, modeling joint multi-object interactions can be intractable due to combinatorial complexity and prohibitive data requirements. 
Instead inspired by~\cite{objectCentricNikhil, sncbf} we adopt an object-centric approach, leveraging a change in input representation to learn a single model that can be scaled to any number of objects. 

The core of our method is a single implicit interaction Control Barrier Function (CBF), $\B$, that reasons about safety from each object's individual perspective.
We achieve this by transforming the system state into the relative coordinate frame for each object. 
Given transformation $R(a, b)$ that maps the state $a$ to the relative coordinates of $b$, we change the model input for object $i$ to be relative to the initial position of object $i$, at time $t-T$:   $(\stateT_{i} = R(\stateT, \objstate^{t-T}_{i}), \objstateHistT_{i} = R(\objstateHistT_{\textbf{abs}}, \objstate^{t-T}_i))$.   
This representation helps to implicitly infer effect of interaction on each object independently: if an object starts moving to the left when pushed forward, this likely means it is interacting with another object that is affecting its motion. 
This object-centric formulation is thus scalable and data-efficient, as a single model trained on representative interactions can generalize to complex scenes with an arbitrary number of objects and novel arrangements.

The trained object-centric model, $\B(\stateT_i, \objstateHistTmone_i) \in \mathbb{R}$, 
predicts that the interaction with the robot at state, $\stateT_i$, given object history $\objstateHistTmone_i$ is safe when $B(\cdot) \geq 0$. 
We compose $\Nobj$ individual object-centric barrier functions into a global scene level CBF that certifies joint safety for all objects: 
\begin{align}
    \label{eq:global_CBF}
    \Bglobal(\stateT_{1:\Nobj}, \objstateHistTmone_{1:\Nobj})  = \min_{i=1, \dots, \Nobj} \B(\stateT_{i}, \objstateHistTmone_{i})
\end{align}
This composition is computationally efficient, scaling linearly with the number of objects, with the individual barrier computation being trivially parallelizable.

\section{Method Implementation: }
In this section we elaborate on the details of how we train and deploy our approach to enable safe, scalable interaction in dense object environments. 
Our method is implemented in two stages: offline learning of the CBF and online deployment as a safety filter. 
The offline process consists of: \textbf{1) Data Collection: } Collecting interaction-rich data in simulation \textbf{2) Initial Training: } Learning an NCBF from the collected data \textbf{3) Refinement: } Iteratively refining the learned CBF to reduce conservativeness. 
The online deployment uses the refined CBF as a safety filter for control synthesis. 
This section describes the details of each process.

\subsection{Online: Safe Set and Safety Filter}


Given the discrete CBF $B$, the induced control invariant set to keep object $i$ safe is $\mathcal{C} = \{(\stateT_i, \objstateHistTmone_i) | \B(\stateT_i, \objstateHistTmone_i) \geq 0\}$. 
Equivalently the global CBF induces a control invariant set to maintain safety with respect to all objects, $\mathcal{C}_{\text{global}} = \{ (\stateTpone_{1:\Nobj}, \objstateHistT_{1:\Nobj}) | \Bglobal(\stateTpone_{1:\Nobj}, \objstateHistT_{1:\Nobj}) \geq 0 \}$. 
At runtime, the global CBF acts as a minimally invasive safety filter. 
Given $\Bglobal$, 
a nominal control, $\action_{\text{nom}}$, and the set of possible robot actions, $\mathbb{U}$, a constrained optimization problem can be solved to generate the minimally invasive safe control $\action_{\text{safe}}$ that satisfies the CBF invariance condition: 
\begin{align}
    \action_{\text{safe}} = \arg\min_{\action \in \mathbb{U}} || \action - \action_{\text{nom}}|| \ s.t. \ \\ \nonumber 
    \Bglobal(f(\stateT_{1:\Nobj}, \action), \objstateHistT_{1:\Nobj}) - \Bglobal(\stateT_{1:\Nobj}, \objstateHistTmone_{1:\Nobj})\\\gamma(\Bglobal(\stateT_{1:\Nobj}, \objstateHistTmone_{1:\Nobj}) ) \geq 0
\end{align}


The above optimization problem may be computationally challenging because we do not assume control-affine dynamics. 
As a practical alternative we employ the following sampling based approach. 
A 
robot's safety can be verified by checking the global CBF value for a potential next state. 
A nominal action, $\action_{\text{nom}}$, is considered safe if it satisfies $\Bglobal(f(\stateT_{1:\Nobj}, \action_{\text{nom}}), \objstateHistT_{1:\Nobj}) \geq 0$. 
If the nominal action is unsafe, we can construct a candidate set of actions, $\mathbb{U}$, by sampling around the nominal control, $\action_{\text{nom}}$. 
From the subset of safe actions, $\mathbb{U}_{\text{safe}} = \{ \action \in \mathbb{U} | \Bglobal(f(\stateT_{1:\Nobj}, \action), \objstateHistT_{1:\Nobj}) \geq 0 \}$, we select the one closest to the nominal control: $\action_{\text{safe}} = \arg\min_{\action \in \mathbb{U}_{\text{safe}}} ||\action - \action_{\text{nom}}||$.
This ensures minimal deviation while maintaining safety. 

If not done in parallel, the CBF computation for every object prior to each action can be computationally expensive. 
In practice, it is often sufficient to only check objects that are either within a fixed distance of the robot or have changed state within their recent history. 
These objects are most relevant for safety because their interactions with the robot, whether direct or through other objects, could critically impact the robot's safety.
This selective evaluation can  reduce computational demands without compromising safety.


\subsection{Offline: Data Collection}
To effectively learn our CBF with implicit interaction modeling, we begin by collecting interaction-rich data in simple simulated environments with up to $\Nobj$ randomly initialized objects. 
Here $\Nobj$ can be much less than the number of objects we want to deploy this model on, as long as it captures most of variations of the robot-object and object-object interactions that will be seen with more objects. 
The dataset must also be representative, using a control policy that captures both safe and unsafe actions to allow the CBF to properly learn the safety boundary. 
This process  produces transition pairs for each object $i$:  $\{(\stateT_i, \objstateHistTmone_i), (\stateTpone_i, \objstateHistT_i)\} ,\forall i \in [1, \dots, \Nobj]$, each labeled as safe or unsafe depending on whether the latter state satisfies the ground-truth safety criterion. 
This forms our safe and unsafe dataset, $\mathcal{D}_{s}$ and $\mathcal{D}_{u}$ respectively.
In addition to the datapoints, we save the corresponding full state of the environment, which is needed for refinement.

\subsection{Offline: Initial Training}
We use a neural network to represent $\B$. 
The network's architecture consists of an LSTM to encode the historical object states and a Multi-Layer Perceptron (MLP) for the robot state. 
These features are fused and passed through a final MLP to output the scalar barrier value.

The initial training optimizes the parameters, $\theta$, of the NCBF by minimizing the loss function $L(\theta)$: 
\begin{align}
        \label{eq:initial_training_loss}
        &L_s(\theta)
        = \frac{1}{|\mathcal{D}_s|}
           \sum_{\mathclap{(\stateT_{i},\,
           \objstateHistTmone_{i})\in\mathcal {D}_s}\vphantom{\Big|}}
           \Big[- \B(\stateT_{i},\,\objstateHistTmone_{i})\Big]_+ ,
        \\
        &L_u(\theta)
        = \frac{1}{|\mathcal{D}_u|}
           \sum_{\mathclap{(\stateT_{i},\,\objstateHistTmone_{i})\in\mathcal{D}_u} \vphantom{\Big|}}
           \Big[\B(\stateT_{i},\,\objstateHistTmone_{i})\Big]_+ ,
    \end{align}
    \begin{align}
        \label{eq:initial_training_loss_pt2}
        &L_d(\theta)
        = \frac{1}{|\mathcal{D}|}
           \sum_{\mathclap{((\stateT_{i},\,\objstateHistTmone_{i}),\,(\stateTpone_{i},\,\objstateHistT_{i}))\in\mathcal{D}} \vphantom{\Big|}}
           \Big[(1-\gamma)\,\B(\stateT_i,\,\objstateHistTmone_i)
                - \B(\stateTpone_{i},\,\objstateHistT_{i}) + \sigma \Big]_+, \\
        &L(\theta) = \eta_{s} \cdot L_{s}(\theta) + \eta_{u} \cdot L_{u}(\theta) + \eta_{d} \cdot L_{d}(\theta).
    \end{align}
Where $[\cdot]_{+}$ is the ReLU function. Here, $L_{s}(\theta)$ enforces that the CBF has positive outputs on safe samples ($\mathcal{D}_s$), $L_{u}(\theta)$ enforces negative outputs for unsafe samples ($\mathcal{D}_u$), and $L_{d}(\theta)$ enforces a penalty for violations of the CBF condition ($\mathcal{D} := \mathcal{D}_s \cup \mathcal{D}_u$). 

\subsection{Offline: Refinement} The CBF trained with this initial policy is often overly conservative, classifying many states as unsafe due to the suboptimal nature of the initial data-collection policy. 
This conservativeness can impede the robot's ability to complete its task. 
Similarly, poor initial training can lead to misclassifications of unsafe states as safe, compromising control invariance of the safe set. 
To address this, we refine the safety  boundary to reduce conservativeness and improve invariance. 
We achieve this by identifying near boundary samples from the dataset where $|\B(\stateT_i, \objstateHistTmone_{i})| \leq \delta$ where $\delta$ is a tunable boundary refinement threshold. 
We then re-simulate these critical scenarios, using the full multi-object scene to capture complex interactions. 
From these states, we apply the ``safest" control, $u_{\text{refinement}}$, which is the action that maximizes the global barrier value:
\begin{align}
    \label{eq:u_refinement}
    u_{\text{refinement}} = \arg\max_{u \in \mathbb{R}^{m}} \Bglobal(f(\stateT_{1:\Nobj}, \action), \objstateHistT_{1:\Nobj})
\end{align}
We take up to $s$ refinement steps, using this safest control strategy and collect all the data pairs, relabeling the states as safe or unsafe, based on the resulting safety in the ground truth simulation. 
This refinement process can be tweaked to solely focus on samples for which $\B(\stateT_i, \objstateHistTmone_{i}) \leq 0$ or $\B(\stateT_i, \objstateHistTmone_{i}) \geq 0$ to make the CBF more or less conservative, respectively. 
The CBF is then retrained on this updated dataset. 
This iterative process can continue until the CBF converges and no misclassifications are found in the dataset, producing our final, control-invariant learned CBF.

\section{Experiments and Results}

We evaluate our method in simulation to demonstrate safe, scalable interaction in cluttered environments where physical contact is unavoidable. 
We designed experiments to assess how performance and generalization change with an increasing number of objects and greater scene complexity.

\textbf{Experiment Setup: }Experiments are conducted in a custom IsaacLab environment containing a Franka Emika Panda arm and several cylindrical bottles shown in Fig.~\ref{fig:globalCBFPlot}(a)
The robot is tasked with reaching a random target location without causing any bottle to tip beyond a $15\degree$ safety threshold. 
The robot's end effector is constrained to a 2D $(x,y)$ planar motion and actions are defined as 2D waypoints with a maximum step size of $1\text{cm}$ while training. 
Each bottle's state is described by its Cartesian coordinates and tipping angle: $(x,y,z,\theta)$, with relative transforms computed by subtracting $(x,y)$ in the cartesian space. 
We test our approach's robustness by randomizing unobservable physical properties for every bottle in every trial, including mass $(1.3 - 2.0 \text{kg})$ and friction coefficients (static: $(0.5-0.7)$, dynamic: $(0.3-0.49)$). 
Observable parameters including the bottle radius $(5\text{cm})$ and height $(20\text{cm})$ were fixed. 
However, observable variations can also be incorporated into our framework through input conditioning.

\begin{figure}[ht]
    \centering
    \begin{subfigure}
        \centering
        \includegraphics[width=0.95\linewidth]{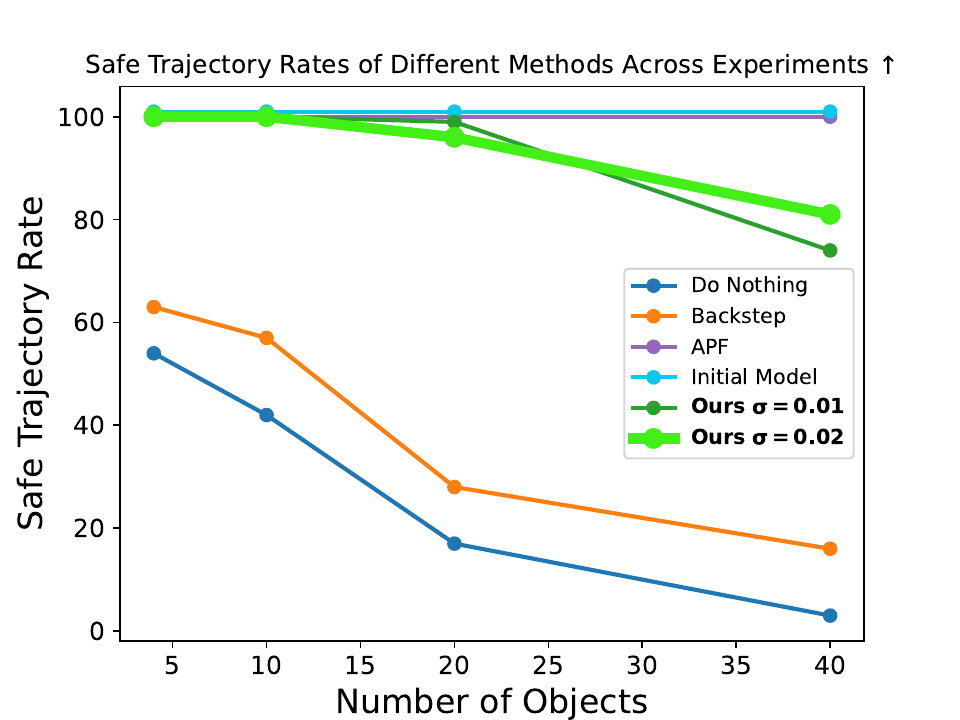}
        \label{fig:linegrapha}
    \end{subfigure}
    \vspace{0.01em}
    \begin{subfigure}
        \centering 
        \includegraphics[clip,width=0.95\linewidth]{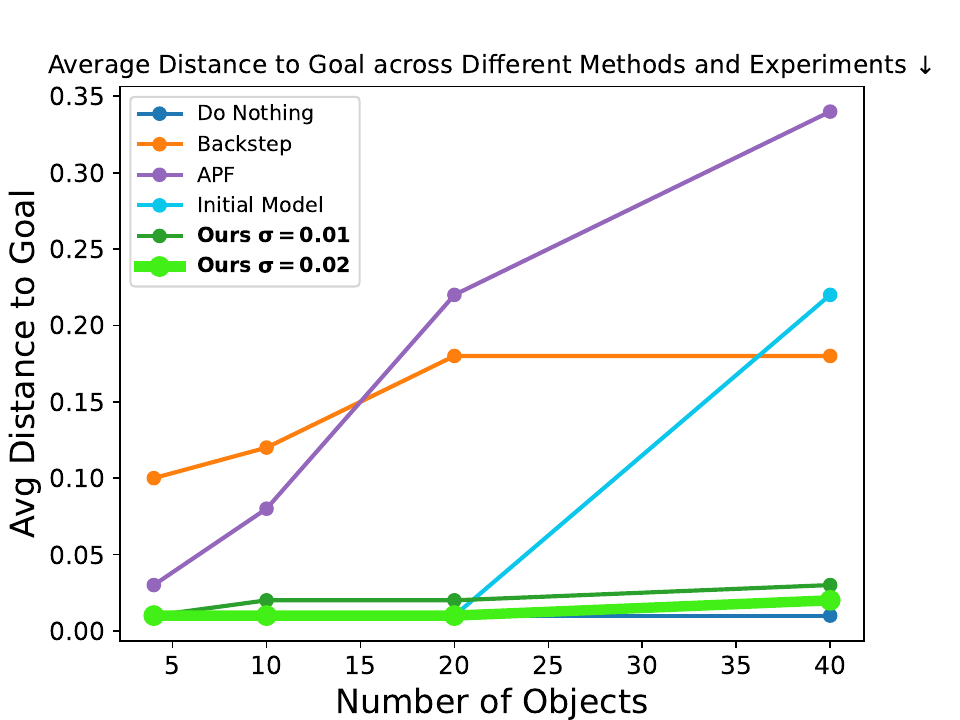}
        \label{fig:linegraphb}
    \end{subfigure}
    \caption{
    Performance of different methods in environments of varying object densities ($4$, $10$, $20$ and $40$ objects).  
    The Safe Rate measures the percentage of safe trajectories (higher is better), and the Average Final Distance to Goal measures final distance to the target (lower is better). 
    These metrics were computed over the same experiment shown in Fig.~\ref{fig:safe_reach_success_rate}, and provide further insight into the performance of each method. 
    These results highlight the limitations of baselines (\emph{Do Nothing}, \emph{Backstepping}) in staying safe, the conservativeness of \emph{APF}, the conservativeness of the \emph{Initial Model} at higher densities, and the robustness of our proposed method, balancing safety and task success.
    }
    \label{fig:linegraphs}
\end{figure}
\textbf{Training and Refinement: } Our CBF is trained in environments containing only $4$ bottles. 
We trained using $\sigma=0.01$ and $\sigma=0.02$. 
We first collected an initial dataset of $9600000$ data pairs from $1200$ trajectories using a naive back-stepping policy. 
This policy steps back to the robot's previous waypoint when nearing the safety threshold.
This dataset was balanced, by discarding samples in free space, to ensure a rich mix of both safe and near-unsafe interaction examples.
To address the limitations of collecting data solely from a
naive controller, specifically the difficulty in obtaining data near the safety boundary, we iteratively refine the model. 
Using IsaacLab's parallel simulation capabilities, we initialized states with high object interaction, collected new data batches with the refinement steps set to $s=4$, and fine-tuned the CBF. 
This process was repeated for $47$ batches. 

\textbf{Visualizing the Learned Safety Boundary: }
To visualize the learned CBF, we generated a global value plot for the $40$-object scene in Fig.~\ref{fig:globalCBFPlot}(b)(c).
The plot shows that the Initial CBF (before refinement) is overly conservative, creating impassable regions that cause the robot to freeze. 
In contrast, the refined CBF learns a tighter safety boundary that identifies safe pathways with physical interaction, enabling successful navigation through the dense environment. 

\textbf{Quantitative Evaluation: }
We evaluated our final model against four baselines in environments with $4,10,20$ and $40$ objects, where success is defined as reaching the goal while maintaining safety. 
The baselines include: \textbf{Do Nothing: } moves straight towards the goal; \textbf{Back-stepping: } the reactive policy used for initial data collection; \textbf{Artificial Potential Fields (APF)}: a classical method using repulsive fields; and \textbf{Initial: } our CBF model before the refinement procedure.
We also compare our method trained with different robustness margins $\sigma=0.01$  and $\sigma=0.02$. 
The Do Nothing Baseline was used as our nominal controller. 
Back-stepping was tuned to step back for any angle approaching $14\degree$, and APF parameters were tuned to $K_P = 5.0$ (attractive potential gain), $\eta = 50.0$ (repulsive potential gain), a potential area with length of $1.2\text{m}$ (size of the table), and an oscillation detection length of $3$, to allow for some contact without being overly conservative.
Fig.~\ref{fig:trajPlot} depicts samples from trajectories using our method (Ours) and the Do Nothing baseline, along with the tilt angles for the highlighted objects of interest. 
The baseline method violates safety for multiple objects, while our method keeps all objects below the safety threshold.

The quantitative results in Fig.~\ref{fig:safe_reach_success_rate} show that our proposed method consistently outperforms all baselines in safety and task success, especially as object density increases.
This metric measures the percentage of trajectories where the robot is both safe and reaches its target.
The results in Fig.~\ref{fig:linegraphs} break down the safe rate and average final distance to goal for each method. 
The naive Do Nothing and Back-stepping approaches fail to adequately consider the effect of robot-object interactions, regularly violating safety and causing objects to tip.
Conversely, APF and our unrefined initial model are overly conservative. 
While safe, they freeze further from the goal in dense scenes, leading to high task failure rate and large final distance from goal. 
Our final refined method strikes a necessary balance, achieving high reach success rates and safety rates when maneuvering through dense object arrangements. 
Comparing $\sigma=0.01$ and $\sigma=0.02$ we see that under-approximating the safety margin can potentially lead to safety violations. 
In principle, an over-approximation could lead to excessive conservativeness. 
Even when our method violates, the average maximum tilt angles in failed trajectories were $17.65\degree$ and $18.88\degree$ for the $20$ and $40$ object cases respectively, showing only minor violations, while the $4$ and $10$ objects cases remained entirely safe (reported for $\sigma=0.02$).
In comparison, the baseline method Do Nothing had much larger violations of $66.81\degree, 68.45\degree, 68.16\degree$ and $46.40\degree$. 
Notably, despite being trained on only $4$ object scenarios, our model generalizes effectively to the complex $40$ object environment, showcasing its scalability and performance.

\section{Conclusion}
In this work, we introduced \ourcbf,  a scalable framework that enables safe robot interaction in densely cluttered environments. 
Our approach successfully generalizes from training on a few objects to complex scenes with many objects, without requiring explicit modeling of intractable multi-object dynamics. 
While our results are promising, several exciting directions for future work remain. 
We plan to validate our method's ability to handle more diverse objects, including those with varying observable parameters. 
Furthermore, we aim to extend our method to more complex scenarios involving unconstrained, full-body robot motion, deformable manipulation and, ultimately, demonstrate its effectiveness on physical hardware. 
This research marks a critical step toward deploying robots that can safely and competently maneuver through the unstructured, contact-rich environments of the real world. 




\bibliographystyle{IEEEtran}
\bibliography{ICRA/example}

@inproceedings{RajaEtAl2024,
  author    = {Raja, S. and Kumar, A. and Gupta, P.},
  title     = {OGM-CBF: Occupancy-Grid Map Based Control Barrier Functions for Safe Robot Control},
  booktitle = {Proc. IEEE Int. Conf. Robot. Autom. (ICRA)},
  pages     = {5678--5685},
  year      = {2024},
}

@misc{TayalEtAl2024,
  author    = {Tayal, M. and Zhang, H. and Jagtap, P. and Clark, A.},
  title     = {Learning a Formally Verified Control Barrier Function in Stochastic Environment},
  year      = {2024},
  eprint    = {2403.19332v1},
  archiveprefix = {arXiv},
  primaryclass = {cs.SY},
}

@ARTICLE{DLOManipulation,
  author={Tang, Yunxi and Chu, Xiangyu and Huang, Jing and Samuel Au, K. W.},
  journal={IEEE Robotics and Automation Letters}, 
  title={Learning-Based MPC With Safety Filter for Constrained Deformable Linear Object Manipulation}, 
  year={2024},
  volume={9},
  number={3},
  pages={2877-2884},
  doi={10.1109/LRA.2024.3362643}}

@unknown{safeClutteredManipulation,
author = {Ding, Xuda and Wang, Han and Ren, Yi and Zheng, Yu and Chen, Cailian and He, Jianping},
year = {2022},
month = {11},
pages = {},
title = {Safety-Critical Optimal Control for Robotic Manipulators in A Cluttered Environment},
doi = {10.48550/arXiv.2211.04944}
}

@inproceedings{HJRSylvia,
author = {Bansal, Somil and Chen, Mo and Herbert, Sylvia and Tomlin, Claire J.},
title = {Hamilton-Jacobi reachability: A brief overview and recent advances},
year = {2017},
publisher = {IEEE Press},
url = {https://doi.org/10.1109/CDC.2017.8263977},
doi = {10.1109/CDC.2017.8263977},
booktitle = {2017 IEEE 56th Annual Conference on Decision and Control (CDC)},
pages = {2242–2253},
numpages = {12},
location = {Melbourne, Australia}
}

@INPROCEEDINGS{CBFAmes,
  author={Ames, Aaron D. and Coogan, Samuel and Egerstedt, Magnus and Notomista, Gennaro and Sreenath, Koushil and Tabuada, Paulo},
  booktitle={2019 18th European Control Conference (ECC)}, 
  title={Control Barrier Functions: Theory and Applications}, 
  year={2019},
  volume={},
  number={},
  pages={3420-3431},
  doi={10.23919/ECC.2019.8796030}}

@article{APFHumanRobotCollab,
  author    = {Dsouza, D. A. and Shenoy, S. and Wang, M. and Chowdhury, A. R.},
  title     = {A comprehensive safety architecture for human--robot collaboration in confined workspaces using improved artificial potential field},
  journal   = {Robotica},
  volume    = {43},
  number    = {4},
  pages     = {1373--1393},
  year      = {2025},
  month     = {apr},
}

@INPROCEEDINGS{APF,
  author={Min Cheol Lee and Min Gyu Park},
  booktitle={Proceedings 2003 IEEE/ASME International Conference on Advanced Intelligent Mechatronics (AIM 2003)}, 
  title={Artificial potential field based path planning for mobile robots using a virtual obstacle concept}, 
  year={2003},
  volume={2},
  number={},
  pages={735-740 vol.2},
  doi={10.1109/AIM.2003.1225434}}

@inproceedings{JIGGLE,
author = {Shinde, Nikhil and Liang, Xiao and Liu, Fei and Zhang, Yutong and Richter, Florian and Herbert, Sylvia and Yip, Michael},
year = {2024},
month = {07},
pages = {},
title = {JIGGLE: An Active Sensing Framework for Boundary Parameters Estimation in Deformable Surgical Environments},
doi = {10.15607/RSS.2024.XX.007}
}

@misc{latentken,
      title={Generalizing Safety Beyond Collision-Avoidance via Latent-Space Reachability Analysis}, 
      author={Kensuke Nakamura and Lasse Peters and Andrea Bajcsy},
      year={2025},
      eprint={2502.00935},
      archivePrefix={arXiv},
      primaryClass={cs.RO},
      url={https://arxiv.org/abs/2502.00935}, 
}

@misc{dohjpponeurips,
      title={Dual-Objective Reinforcement Learning with Novel Hamilton-Jacobi-Bellman Formulations}, 
      author={William Sharpless and Dylan Hirsch and Sander Tonkens and Nikhil Shinde and Sylvia Herbert},
      year={2025},
      eprint={2506.16016},
      archivePrefix={arXiv},
      primaryClass={cs.AI},
      url={https://arxiv.org/abs/2506.16016}, 
}

@misc{oswinneurips,
      title={Solving Minimum-Cost Reach Avoid using Reinforcement Learning}, 
      author={Oswin So and Cheng Ge and Chuchu Fan},
      year={2024},
      eprint={2410.22600},
      archivePrefix={arXiv},
      primaryClass={cs.LG},
      url={https://arxiv.org/abs/2410.22600}, 
}

@InProceedings{backtobase,
  title = 	 {Back to Base: Towards Hands-Off Learning via Safe Resets with Reach-Avoid Safety Filters},
  author =       {Begzadic, Azra and Shinde, Nikhil and Tonkens, Sander and Hirsch, Dylan and Ugalde, Kaleb and Yip, Michael and Cortes, Jorge and Herbert, Sylvia},
  booktitle = 	 {Proceedings of the 7th Annual Learning for Dynamics \&amp; Control Conference},
  pages = 	 {1154--1166},
  year = 	 {2025},
  editor = 	 {Ozay, Necmiye and Balzano, Laura and Panagou, Dimitra and Abate, Alessandro},
  volume = 	 {283},
  series = 	 {Proceedings of Machine Learning Research},
  month = 	 {04--06 Jun},
  publisher =    {PMLR},
  url = 	 {https://proceedings.mlr.press/v283/begzadic25a.html}
}

@misc{deepreach,
      title={DeepReach: A Deep Learning Approach to High-Dimensional Reachability}, 
      author={Somil Bansal and Claire Tomlin},
      year={2020},
      eprint={2011.02082},
      archivePrefix={arXiv},
      primaryClass={cs.RO},
      url={https://arxiv.org/abs/2011.02082}, 
}

@misc{deepreachmpc,
      title={Bridging Model Predictive Control and Deep Learning for Scalable Reachability Analysis}, 
      author={Zeyuan Feng and Le Qiu and Somil Bansal},
      year={2025},
      eprint={2505.03830},
      archivePrefix={arXiv},
      primaryClass={cs.RO},
      url={https://arxiv.org/abs/2505.03830}, 
}

@misc{forcefeedback,
  author    = {Wei, L. and Ma, J. and Hu, Y. and Zhang, R.},
  title     = {Ensuring Force Safety in Vision-Guided Robotic Manipulation via Implicit Tactile Calibration},
  year      = {2024},
  eprint    = {2412.10349v1},
  archiveprefix = {arXiv},
  primaryclass = {cs.RO},
}

@misc{shortHorizonMPC,
  author    = {Lee, J. and Seo, M. and Bylard, A. and Sun, R. and Sentis, L.},
  title     = {Real-Time Model Predictive Control for Industrial Manipulators with Singularity-Tolerant Hierarchical Task Control},
  year      = {2022},
  eprint    = {2209.11880},
  archiveprefix = {arXiv},
  primaryclass = {cs.RO},
}

@misc{cpo,
      title={Constrained Policy Optimization}, 
      author={Joshua Achiam and David Held and Aviv Tamar and Pieter Abbeel},
      year={2017},
      eprint={1705.10528},
      archivePrefix={arXiv},
      primaryClass={cs.LG},
      url={https://arxiv.org/abs/1705.10528}, 
}

@misc{gu2025balancerewardsafetyoptimization,
      title={Balance Reward and Safety Optimization for Safe Reinforcement Learning: A Perspective of Gradient Manipulation}, 
      author={Shangding Gu and Bilgehan Sel and Yuhao Ding and Lu Wang and Qingwei Lin and Ming Jin and Alois Knoll},
      year={2025},
      eprint={2405.01677},
      archivePrefix={arXiv},
      primaryClass={cs.LG},
      url={https://arxiv.org/abs/2405.01677}, 
}

@inproceedings{macklin2016xpbd,
  title={XPBD: position-based simulation of compliant constrained dynamics},
  author={Macklin, Miles and M{\"u}ller, Matthias and Chentanez, Nuttapong},
  booktitle={Proceedings of the 9th International Conference on Motion in Games},
  pages={49--54},
  year={2016}
}

@article{isaaclab,
   author={Mittal, Mayank and Yu, Calvin and Yu, Qinxi and Liu, Jingzhou and Rudin, Nikita and Hoeller, David and Yuan, Jia Lin and Singh, Ritvik and Guo, Yunrong and Mazhar, Hammad and Mandlekar, Ajay and Babich, Buck and State, Gavriel and Hutter, Marco and Garg, Animesh},
   journal={IEEE Robotics and Automation Letters},
   title={Orbit: A Unified Simulation Framework for Interactive Robot Learning Environments},
   year={2023},
   volume={8},
   number={6},
   pages={3740-3747},
   doi={10.1109/LRA.2023.3270034}
}

@article{discreteCBF,
   title={Counterexample-Guided Synthesis of Robust Discrete-Time Control Barrier Functions},
   volume={9},
   ISSN={2475-1456},
   url={http://dx.doi.org/10.1109/LCSYS.2025.3578971},
   DOI={10.1109/lcsys.2025.3578971},
   journal={IEEE Control Systems Letters},
   publisher={Institute of Electrical and Electronics Engineers (IEEE)},
   author={Shakhesi, Erfan and Katriniok, Alexander and Heemels, W. P. M. H. Maurice},
   year={2025},
   pages={1574–1579} }

@ARTICLE{parameterizedCBF,
  author={Alan, Anil and Molnar, Tamas G. and Ames, Aaron D. and Orosz, Gábor},
  journal={IEEE Control Systems Letters}, 
  title={Parameterized Barrier Functions to Guarantee Safety Under Uncertainty}, 
  year={2023},
  volume={7},
  number={},
  pages={2077-2082},
  doi={10.1109/LCSYS.2023.3285188}}

@misc{ncbf,
      title={Safe Control Under Input Limits with Neural Control Barrier Functions}, 
      author={Simin Liu and Changliu Liu and John Dolan},
      year={2022},
      eprint={2211.11056},
      archivePrefix={arXiv},
      primaryClass={cs.RO},
      url={https://arxiv.org/abs/2211.11056}, 
}

@misc{refinecbf,
      title={Refining Control Barrier Functions through Hamilton-Jacobi Reachability}, 
      author={Sander Tonkens and Sylvia Herbert},
      year={2022},
      eprint={2204.12507},
      archivePrefix={arXiv},
      primaryClass={cs.RO},
      url={https://arxiv.org/abs/2204.12507}, 
}

@misc{sncbf,
      title={Sequential Neural Barriers for Scalable Dynamic Obstacle Avoidance}, 
      author={Hongzhan Yu and Chiaki Hirayama and Chenning Yu and Sylvia Herbert and Sicun Gao},
      year={2023},
      eprint={2307.03015},
      archivePrefix={arXiv},
      primaryClass={cs.RO},
      url={https://arxiv.org/abs/2307.03015}, 
}

@misc{objectCentricNikhil,
      title={Object-centric Representations for Interactive Online Learning with Non-Parametric Methods}, 
      author={Nikhil U. Shinde and Jacob Johnson and Sylvia Herbert and Michael C. Yip},
      year={2023},
      eprint={2307.10063},
      archivePrefix={arXiv},
      primaryClass={cs.RO},
      url={https://arxiv.org/abs/2307.10063}, 
}

@misc{marcosafemanipulation,
      title={Safe, Task-Consistent Manipulation with Operational Space Control Barrier Functions}, 
      author={Daniel Morton and Marco Pavone},
      year={2025},
      eprint={2503.06736},
      archivePrefix={arXiv},
      primaryClass={cs.RO},
      url={https://arxiv.org/abs/2503.06736}, 
}

@misc{cbfchallenges,
      title={Advances in the Theory of Control Barrier Functions: Addressing Practical Challenges in Safe Control Synthesis for Autonomous and Robotic Systems}, 
      author={Kunal Garg and James Usevitch and Joseph Breeden and Mitchell Black and Devansh Agrawal and Hardik Parwana and Dimitra Panagou},
      year={2023},
      eprint={2312.16719},
      archivePrefix={arXiv},
      primaryClass={math.OC},
      url={https://arxiv.org/abs/2312.16719}, 
}

\end{document}